\documentclass[11pt]{article}

\usepackage[final]{acl}

\usepackage{times}
\usepackage{latexsym}

\usepackage[T1]{fontenc}

\usepackage[utf8]{inputenc}

\usepackage{microtype}

\usepackage{inconsolata}

\usepackage{graphicx}
\usepackage{amsmath}
\usepackage{amssymb}
\usepackage{color}
\usepackage{booktabs}
\usepackage{multirow}
\usepackage{adjustbox}
\usepackage{caption}

\usepackage{mathtools}

\usepackage[linesnumbered,ruled,vlined]{algorithm2e}

\newcommand{\revise}[1]{\textcolor{black}{#1}}

\newcommand{\eat}[1]{}

\title{CAST: Character-and-Scene Episodic Memory for Agents}

\author{Kexin Ma\thanks{These authors contribute equally to this work.}, 
Bojun Li\footnotemark[1], \\
Yuhua Tang, 
Liting Sun,
Ruochun Jin\thanks{The corresponding author.}
\\
College of Computer Science and Technology, \\
National University of Defense Technology, Changsha, China \\
\small{
   \textbf{Correspondence:} \href{mailto:email@domain}{\{makexin, jinrc\}@nudt.edu.cn}  
   }
}

\begin{document}
\maketitle
\begin{abstract}
Episodic memory is a central component of human memory, which refers to the ability to recall coherent events grounded in \emph{who}, \emph{when}, and \emph{where}.
However, most agent memory systems only emphasize semantic recall and treat experience as structures such as key-value, vector, or graph,
which makes them struggle to represent and retrieve coherent events.
To address this challenge, we propose a Character-and-Scene
based memory architecture(\textbf{CAST}) inspired by dramatic theory. 
Specifically, CAST constructs 3D scenes (time/place/topic) and organizes them into character profiles that summarize the events of a character to represent episodic memory. 
Moreover, CAST complements this episodic memory with a graph-based semantic memory, which yields a robust dual memory design.
Experiments demonstrate that CAST has averagely improved 8.11\% F1 and 10.21\% J(LLM-as-a-Judge) than baselines on various datasets, especially on open and time-sensitive conversational questions\footnote{Our code are available at \url{https://anonymous.4open.science/r/code-for-paper-0CB7/}}.

\end{abstract}

\section{Introduction}
\revise{Human memory is not a flat key–value store. 
	Cognitive science has proposed that
	humans maintain multiple, structurally distinct memory systems including 
	semantic memory for abstract facts and concepts,
	and episodic memory for concrete events situated in time, place, and social context~\cite{tulving1972episodic,schacter1994memory}.
	When recalling a person Gina,
	one rarely retrieves an isolated proposition, e.g. ``Gina likes contemporary dance``.
	Instead, one may reconstruct a moment that ``in that evening at the studio,
	Gina’s team won the first place with the contemporary piece`` \cite{maharana2024evaluating},
	connect it to the memory of Gina and conclude she likes contemporary dance.
	Notably, such moment reconstruction has been
	implemented as concepts of \textit{scene} and \textit{character profiles} in dramatic theory~\cite{richards1895butcher,corrigan2012film,egri1972art}.
	Specifically, a scene requires the consistency of time, places and actions,
	which comes from \textit{classical unities}~\cite{richards1895butcher,corrigan2012film}, a principle of dramatic structure.
	Character profiles summarize how character’s traits and relationships unfold across scenes~\cite{egri1972art}.
	Together, these perspectives suggest a potential episodic memory for agents,
	which not only store plain facts,
	but also what happened to whom, when, and where.}

\begin{figure*}[!t]
	\centering 
	\includegraphics[width=\linewidth]{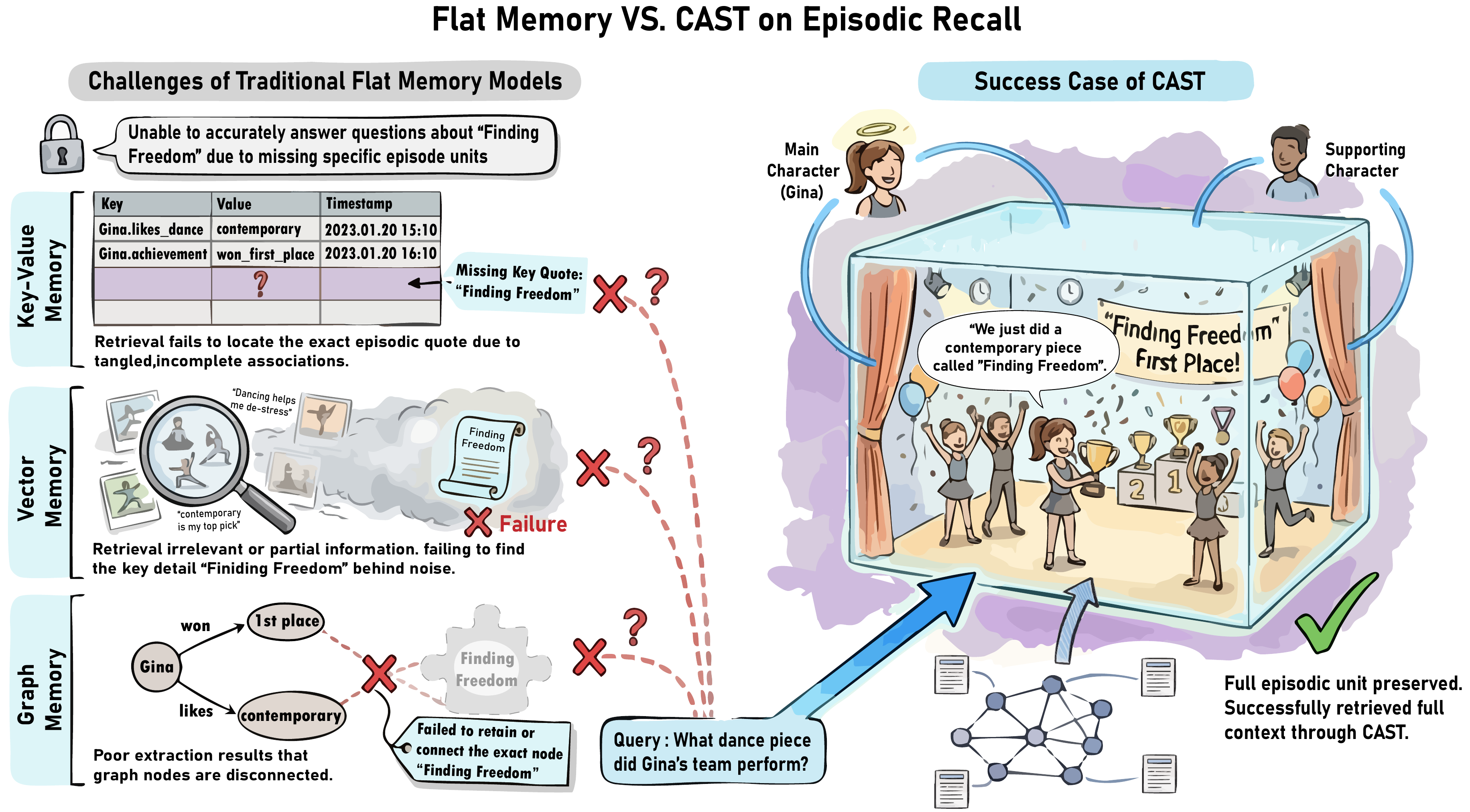}
	\caption{{Comparison between the flat memory of current LLM agents (left) and our CAST inspired by human cognition and dramatic theory (right)}.}
	\label{fig:intro}
\end{figure*}

However, large language models (LLMs) and their memory management fail to support faithful episodic recall, where even the most advanced LLMs struggle with episodic memory tasks\cite{2025epmembench}.
Existing agent memory management mainly falls into three paradigms\cite{hu2025memory}, i.e., key-value (KV), vector-based and graph memories.
KV memories store experiences as key–value pairs (e.g., $\text{Gina.likes\_dance = contemporary}$)\cite{weston2014memory,zhang2017dynamic}. 
While efficient, this approach is sensitive to the extraction of keys and values,
and lacks semantic connections between related episodes\cite{shuster2020deploying}.
As in Figure~\ref{fig:intro}, since there are no keywords ``dance piece`` and relevant semantic connections,
the KV memory fails to find the correct answer ``finding freedom``.
Vector-based memories embed all content into a shared similarity space\cite{zhong2024memorybank,han2025legomem,lee2023prompted}. 
Nevertheless, vector retrieval via approximate nearest neighbors often confuses similarity with equality{\cite{ma2024context,YaoMJZW25}}, which makes it difficult to find accurate relevant documents.
For example, vector memory retrieval may return a generic mention ``I love dancing`` instead of the correct answer ``finding freedom`` in Figure~\ref{fig:intro}.
Graph memories are skilled at discovering connections between sentences while relying heavily on the quality of knowledge extraction\cite{komeili2022internet,petroni2019language}. 
Worse yet, informal references and implicit context are frequently omitted in open-domain dialogue,
which fragments the graph and breaks critical links necessary for precise recall.
In Figure~\ref{fig:intro}, due to the lack of explicit mention of ``dance piece`` in the original text and the rough extraction of nodes, ``finding freedom`` is ignored.

Inspired by and based on the key concepts in dramatic theory,
we propose \textbf{C}har\textbf{a}cter-and-\textbf{S}cene based memory archi\textbf{t}ecture (\textbf{CAST}),
which is organized by character profiles for agents
and combines semantic memory and episodic memory to address these limitations. 
Specifically, CAST constructs a graph-based index for semantic memory and a scene-based index for episodic memory.
For semantic memory, a heterogeneous graph is built with edges capturing semantic similarity.
For episodic memory, we define each {message} in dialogue as {a 3D view with time, place, and action features},
and aggregate multiple views into scenes when {they are close 3D features},
thereby {following} the {classical unities} in a 3D episodic index. 
{Furthermore, CAST extracts all the involved participants from views, and annotate their roles, main characters and supporting characters.
Finally the corresponding scenes are added to the specific character profiles.}
That is, these character-centric episodes act as character profiles that summarize ``what happened to whom, when, and where``. 
During retrieval, the agent queries both semantic and scene memory, where fusion strategies are designed for an appropriate combination of memories.
As a result, CAST preserves full episodic context, which avoids the {rigidity} of KV, the ambiguity of vector search, and the fragility of graph extraction, and supports accurate answers to fine-grained episodic questions.

We have evaluated CAST on the LOCOMO\cite{maharana2024evaluating} and epbench\cite{2025epmembench} benchmarks.
Experiments show that CAST outperforms naive RAG on average by 26.54\% in F1 and 32.12\% in J(LLM-as-a-Judge), and exceeds all the baselines 8.11\% F1 and 10.21\% J averagely.
Among them, enhancement of F1 and J can reach up to 24.25\% and 46.59\%, repectively.

\section{Related Work}
\paragraph{Agent Memory for LLM-Based Systems.}
Researchers have augmented
LLM agents with long-term memory.
The simplest approach relies on long contexts, where agents maintain a rolling window of recent dialogues\cite{zhang2018personalizing}.
While effective for very short interactions, this strategy scales poorly with interaction length\cite{maharana2024evaluating}.
A second line of work provides explicit memory modules as vector stores\cite{lewis2020retrieval,guu2020realm,izacard2021leveraging}.
Agent memory plugins such as MemoryBank\cite{zhong2024memorybank}, MemGPT\cite{beurer2023memgpt} embed data into a high-dimensional space and store it with optional metadata (timestamps, user IDs, tags).
More recent research begins to introduce explicit structure and memory management\cite{beurer2023memgpt,park2023generative}.
Systems such as Zep\cite{rasmussen2025Zep}, Mem0\cite{chhikara2025Mem0}, 
organize and summary memories and use scoring to decide which items to retain or compress.
HippoRAG\cite{jimenez2024hipporag,gutierrez2025rag} and Mirix\cite{wang2025mirix} have preliminarily organized structural memory, while leave the exploration of episodic memory remains to be done.
Different from previous work, which stores contents as flat entries, our work factors originally scattered fragments into units tied to specific characters and scenes,
such that retrieval better captures contextual threads and implicit cues.

\paragraph{Story Organization.}
Events~\cite{chambers2008unsupervised}, scripts~\cite{schank1977scripts} and scenes~\cite{hearst1997texttiling} are three common units to organize events of stories from raw text.
Among them, scenes are utilized in multiple domains.
\citet{hearst1997texttiling} explicitly models scenes by segmenting continuous text into coherent stretches that act as local story units. 
{3D computer vision similarly treats a scene as a
spatially coherent environment recovered from multiple views via multi-view feature aggregation\cite{seitz2006comparison,sitzmann2019scene,mildenhall2020nerf},
which reinforces the insight of a scene as a unit that integrates many local observations into a single structured representation.}
Moreover, narrative understanding and dialogue systems {need to reason about people, not just events}\cite{bruner1991narrative,liu2020character}.
In NLP, character and persona modeling typically encodes who a character is, rather than what has happened to them across time and space. 
\citet{bamman2013learning} represent each character as a mixture of prototypical roles and attributes.
Generative agent frameworks~\cite{park2023generative} and structured memory systems allow agents to store observations about others. 
Cognitive theories of human memory suggest a richer picture of character profiles. 
Autobiographical memory research~\cite{tulving1972episodic,conway2000construction,schacter1998cognitive} argues that personal and social memories are organized hierarchically into lifetime periods, general events, and specific episodes, often indexed by cues based on person, time, and place. 
Our work brings these perspectives together and adapts to agent memory,
which treats character profiles as the top layer retrieval units in an LLM agent’s memory,
leveraging the dramatic-inspired structure as functional components.

\paragraph{Cognitive Foundations of Agent Memory.}
Human episodic memory (EM) is structured around discrete events, not continuous streams. 
Cognitive science shows that people segment experience at boundaries where goals, locations, or actions shift\cite{zacks2007event}, which enables efficient encoding and retrieval.
Early in 2010s, researchers \citet{brom2010episodic} have begun to find the ability to learning from human memory to design agent memory.
For instance, ~\citet{fountas2025human} have identified ``surprise-driven'' event boundaries in long text streams and constructed an episodic index over segmented events for LLMs.
~\citet{wu2025human} then proposed several key design principles for building human-like episodic memory systems in LLMs.
Different from prior work that infers event boundaries from prediction errors,
CAST is inspired by dramatic theory, utilizing scenes that enable rapid and interpretable segmentation of experience into event structures, without relying on fine-grained internal model signals.

\section{Character-and-scene based Memory Architecture}

\subsection{Preliminaries}

We assume a multi-turn dialogue
\(
\mathcal{C} = (m_1, m_2, \dots, m_T),
\)
where each message
\(
m_i = (\text{speaker}_i, t_i, \ell_i, x_i)
\)
consists of a speaker identifier, timestamp, location label, and text.

\paragraph{View.}
We take a short dialogue window as the basic observation unit, view. 
For \(i \in \{1,\dots,T\}\), we form a  length $w$ window
\(
v_{i_w} = (m_{i-w},..., m_i,..., m_{i+w}),
\)
where if $i-w\le 1$,$\quad m_{i-w} \coloneqq m_{i-w+1}$,
if  $i+w\ge T$,$\quad m_{i+w} \coloneqq m_{i+w-1}$.
Taking both efficiency and accuracy into account, we set $w = 1$(details can be seen in Section~\ref{sec:analysis}), 
{in which case the view is written simply as $v_i$.}
Let
\(
\mathcal{V} = \{v_1,\dots,v_T\}
\)
denote the set of all views. 
{For any \(v \in \mathcal{V}\), we write
\(
v = <v.x,\, v.t,\, v.\ell,\, v.P>,
\)
where \(v.x\) is the concatenated text of messages, \(v.t\) and \(v.\ell\) are taken from the central message, and \(v.P\) is the set of participants in the window.} 
In addition, $v.\tau$ is denoted as the topic of $v$, which is extracted from $v.x$.


\paragraph{Scene.}
A scene is defined as a set of views that are close in time, location and topic. 
Formally, a scene is
a non-empty subset \(s \subseteq \mathcal{V}\) such {that for all \(v, v' \in s\),
	\(
	|v.t - v'.t| \le \Delta_t,\quad d_\ell(v.\ell, v'.\ell) \le \Delta_\ell,\quad d_\tau(v.\tau, v'.\tau) \le \Delta_\tau,
	\)}
where \(d_\ell\) and \(d_\tau\) are distance functions in location and topic spaces, and \(\Delta_t, \Delta_\ell, \Delta_\tau\) are preset thresholds. 
For any scene \(s\), we define
its participant set
\(
P(s) = \bigcup_{v\in s} v.P,
\)
and a scene summary \(\phi(s)\) to describe the scene, which are embedded into {vector}  \(\mathbf{h}_s\).
The set of scene vectors \(\mathbf{h}_s\) is denoted as $H$.

\paragraph{Main and supporting characters.}  
Within a scene, participants are divided into main and supporting characters.
Main characters (MC) are those who play significant roles in the scene, while supporting characters (SC) are usually of minor importance.
{Formally}, for any scene \(s\) with participant set \({P(s)}\) and any character \(c \in P(s)\), we define a role label
$
\rho(c,s) = \text{MC}, \text{if } c\ \text{is the main character.} 
$
Otherwise, $\rho(c,s) = \text{SC}.$
\paragraph{Character profile.}
A character profile summarizes how a person traverses scenes over time, annotated with their role in each scene. 
Let $\mathcal{S} = \allowbreak \{ s_1,\allowbreak\dots,\allowbreak s_K\}$ denote the set of all scenes constructed from the dialogue.
For character \(c\), the scenes where \(c\) appears
are collected 
$\mathcal{S}(c) \allowbreak = \{\, s \allowbreak  \in  \mathcal{S} \allowbreak : c \in P(s) \allowbreak \,\}$
and sorted by start time, which is denoted by
\(\mathcal{S}(c) = \{s_{c,1},\dots,s_{c,K_c}\}\).
Then the character profile of \(c\) is
\(
\pi(c) = \big( (s_{c,1}, \rho(c,s_{c,1})), \dots, (s_{c,K_c}, \rho(c,s_{c,K_c})) \big),
\)

\begin{figure}[!t]
	\centering 
	\includegraphics[width=\columnwidth]{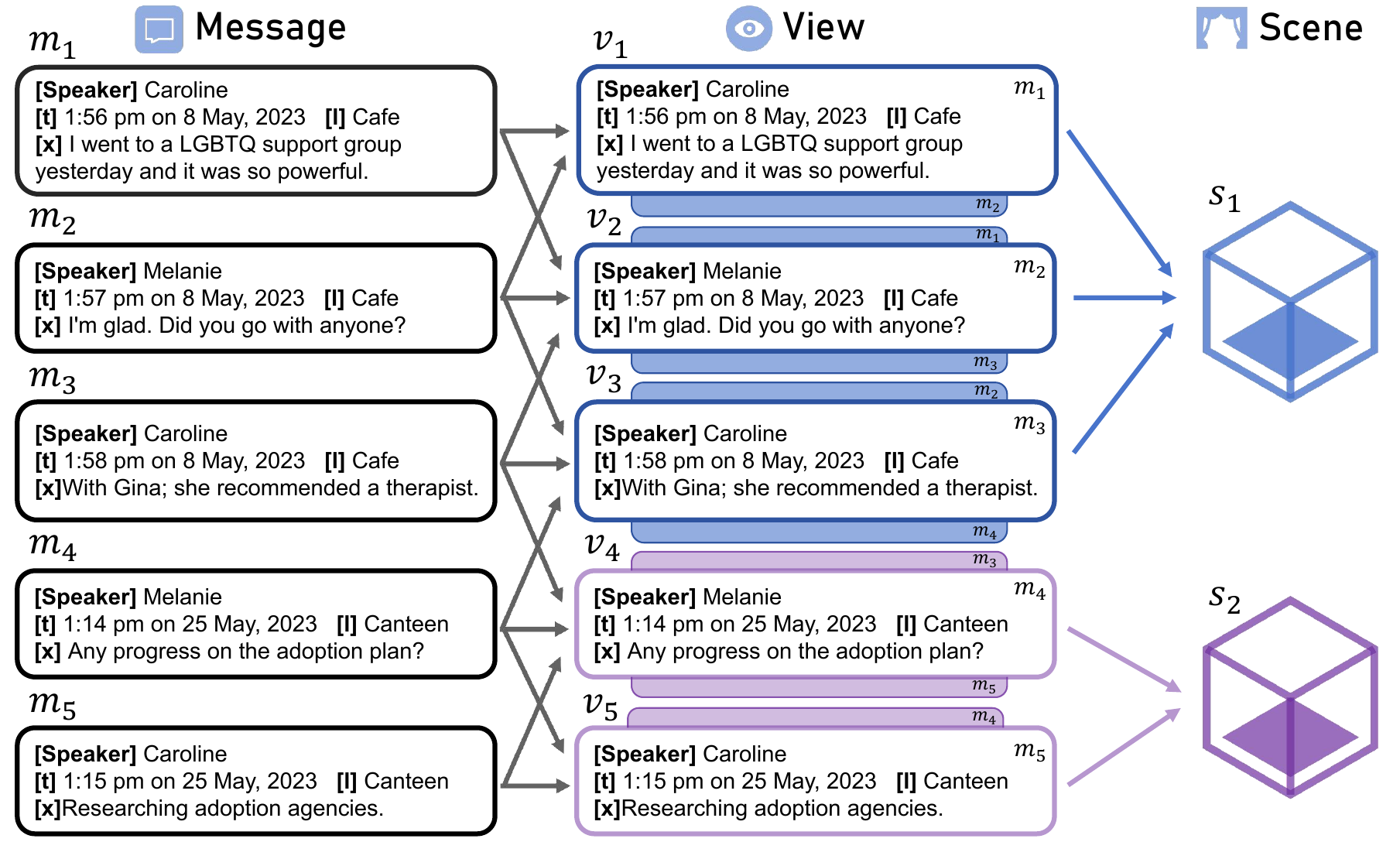}
	\caption{Example of scene aggregation.}
	\label{fig:ex}
\end{figure}

\paragraph{Examples.}
Consider the messages $\{m_1, \allowbreak\dots\allowbreak,m_5\}$ in Figure~\ref{fig:ex}.
With window size $w\allowbreak=\allowbreak1$, the view centered at $m_1$ is $v_1\allowbreak=\allowbreak\{m_1\allowbreak,m_2\}$ and centered at $m_2$ is $v_2\allowbreak=\{m1\allowbreak,m2\allowbreak,m3\}$, whose participant sets includes $\{\text{Caroline, Melanie}\}$ and $\{\text{Caroline, Melanie, Gina}\}$, respectively. 
CAST then finds that $v_1,v_2,v_3$ are close in time and place, which forms a scene $s_1$.
Meanwhile, $v_4,v_5$ occur later and form a separate scene $s_2$.
In $v_2$, Gina is mentioned but not speaking, labeled with $\text{SC}$, 
while Caroline and Melanie are speakers in their respective turns, labeled with $\text{MC}$.

\subsection{CAST Overview}

\begin{figure*}[!t]
	\centering 
	\includegraphics[width=\linewidth]{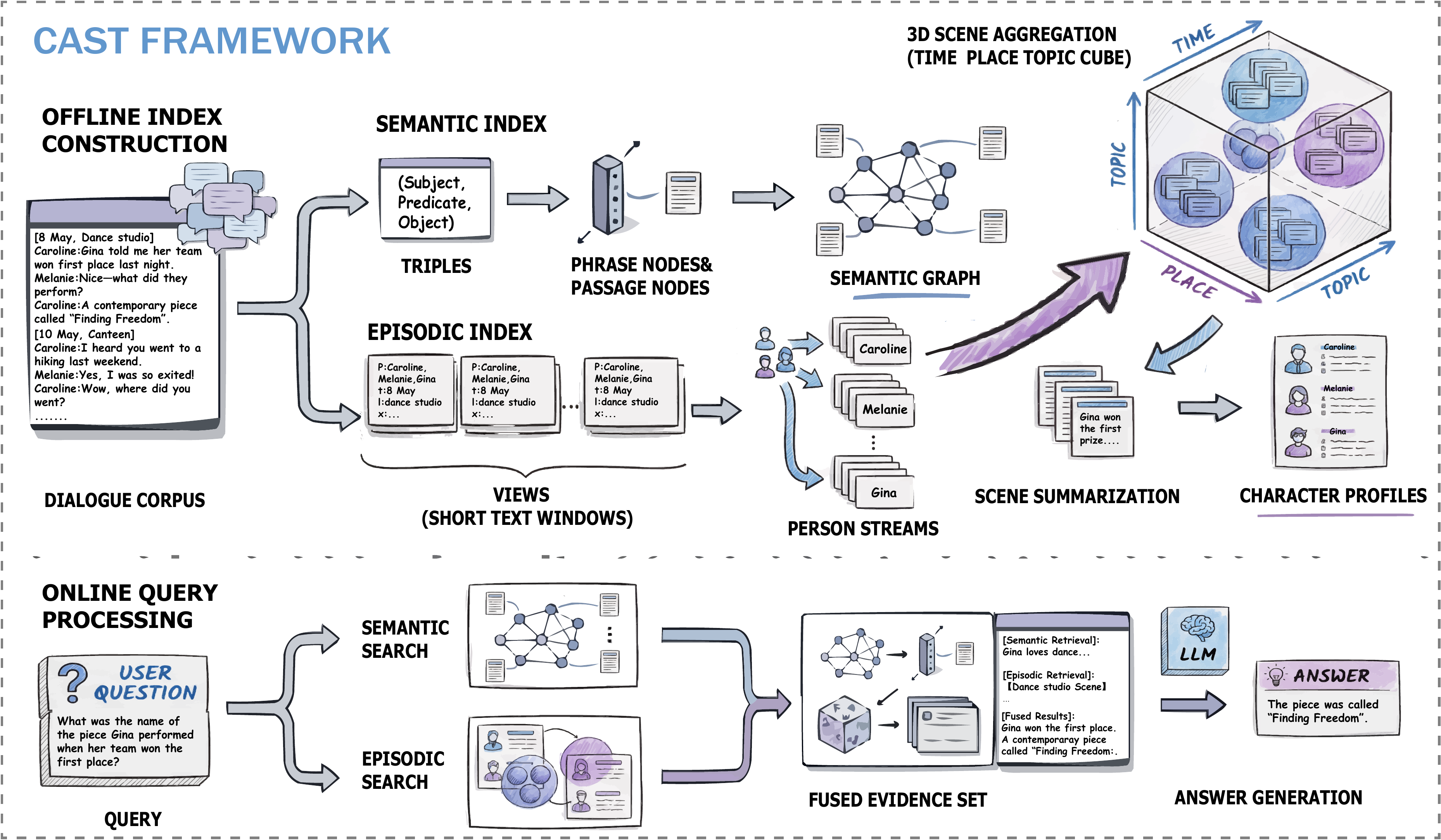}
	\caption{The overview of CAST.}
	\label{fig:overview}
\end{figure*}

CAST is designed for agents with concepts borrowed from dramatic theory,
which models human-like memory by semantic and episodic memory. 
Overall, the framework consists of two stages, an offline index construction stage and online query processing stage as shown in Figure~\ref{fig:overview}. 
Below is a brief overview of CAST, more crucial details are explained in Section~\ref{sec:detailtec}.

\paragraph{Offline Index Construction.}
The offline stage takes as input a dialogue corpus and produces two persistent indices: a semantic index and an episodic index.
The semantic index is a hybrid indexing architecture that follows HippoRAG2\cite{gutierrez2025rag}, integrating a graph memory with Dense Passage Retrieval (DPR)\cite{karpukhin2020dense}. 
CAST first {utilizes} an LLM with OpenIE\cite{angeli-etal-2015-leveraging} to extract (subject, predicate, object) triples, then form phrase and passage nodes, finally construct a semantic graph.
Meanwhile, the episodic index is constructed through three main stages.
First, the raw dialogue is converted into views, optionally using a fixed adjacent window to capture local context.
Then, for each view, CAST identify the speaker and all the mentioned persons, and replicate the view into each identified person’s stream with role labels.
Finally, within each person stream, a greedy 3D clustering by time, place and topic is performed to obtain scenes of the character, and summarize the scene set as a character profile.

\paragraph{Online Query Processing.}
Given a question \(q\), the system retrieves evidence from both the semantic graph and the episodic index, and then fuses the results for answer generation.
Specifically, 
the semantic retrieval follows HippoRAG2, which first searches in the semantic graph. 
Meanwhile, DPR is performed in parallel. 
If the graph retrieval yields no results, the system falls back to the DPR search results.
The semantic retrieval result is denoted as \(\mathcal{E}_{\text{sem}}(q)\), which is particularly effective for multi-hop factual questions.
For the episodic index, we encode \(q\) using the scene encoder to obtain \(\mathbf{z}_q = f_{\text{scene}}(q)\) and perform \(k\)-nearest-neighbor search over the scene embeddings \(H\), retrieving scenes whose summaries are semantically similar to the question. 
Named entities, time expressions, and locations from \(q\) are then extracted and used to filter scenes. 
\eat{For questions that mention specific characters, we restrict candidates to scenes drawn from the corresponding profiles \(\bigcup_c \mathcal{S}(c)\), focusing episodic retrieval on the relevant storylines. }
Retrieved scenes are mapped back to their underlying views and messages, yielding an episodic retrieval result \(\mathcal{E}_{\text{epi}}(q)\) that targets questions about what happened, when, where, and to whom.
Finally, we combine \(\mathcal{E}_{\text{sem}}(q)\) and \(\mathcal{E}_{\text{epi}}(q)\) into a single evidence set \(\mathcal{E}(q)\), which are provided to a language model to generate answers. 
\eat{In this way, the semantic index and episodic index jointly support retrieval, mirroring the division between semantic and episodic memory in human cognition.}

\subsection{Detailed Methodology}
\label{sec:detailtec}
\subsubsection{Role assignment}
\label{sec:ra}
Main characters are protagonists whose intentions drive the plot, while supporting characters are those who react, assist, or provide contexts\cite{aristotle1996poetics}. 
This role hierarchy enables audiences to prioritize attention and maintain coherent mental models of ongoing narratives{\cite{lessing1879hamburgische}}.
Inspired by this principle, we assign role labels within each scene to support memory organization with different levels of attention.

Given a set of views ${v_i}$ segmented from the dialogue stream, CAST first associates each view with its corresponding speaker, 
e.g., $v_1$ is linked to the speaker of $m_1$. 
As a result, views are grouped by speakers to form individual character streams.
Then for each character stream, we perform named entity recognition (NER) to extract all person names mentioned within its constituent views. 
This yields a comprehensive list of participants across the scene.
Next, role labels are assigned based on the principles below.
Any participant who serves as the speaker of at least one view in the scene is designated as a main character ($ \rho(c,s)= \text{MC} $), reflecting their active involvement in the {unfolding} event.
All the other identified participants, those who are only referenced but never speak within the scene, are labeled as supporting characters ($ \rho(c,s)= \text{SC} $).
Notably, a character may be an $\text{MC}$ in one scene and an $\text{SC}$ in another, which enables fine-grained profiling.

\subsubsection{Scene aggregation}
Coherent structure emerges from aggregating complementary observations.
In narrative understanding, a single event is often observed through multiple utterances,  
which mirrors the idea in 3D scene reconstruction\cite{seitz2006comparison,sitzmann2019scene}. 
Drawing on this intuition, we proposed our scene aggregation, which performs a greedy 3D clustering by time, place and topic.

Views are processed to character streams and labeled in Section~\ref{sec:ra}.
A current scene \(s\) is maintained as a growing set of views, 
initially \(s = \{v_1\}\).
For each subsequent view \(v_i\) we compute its temporal, spatial, and topical distances to the current scene:
\(
\delta_t = |v_i.t - t_{\max}(s)|,\quad
\delta_\ell = d_\ell(v_i.\ell, \ell(s)),\quad
\delta_\tau = d_\tau(v_i.\tau, \tau(s)),
\)
where \(\ell(s)\) and \(\tau(s)\) are running representatives (e.g., centroids) of location and topic within \(s\). 
If
two of the three conditions
\(
\delta_t \le \Delta_t,\quad \delta_\ell \le \Delta_\ell,\quad \delta_\tau \le \Delta_\tau,
\)
holds,
CAST add \(v_i\) to \(s\) and update its statistics.
Otherwise a new scene is started with \(v_i\). 
This procedure yields a partition of the interaction stream into non-overlapping scenes $\mathcal{S}=\{s_1,…,s_K\}$.
Finally, {a scene representation \(\phi(s)\) is constructed by summarizing} the texts \(\{v.x: v \in s\}\) for each scene $s$. 
Then a pretrained encoder \(f_{\text{scene}}\) is utilized to obtain a scene summary vector
\(
\mathbf{h}_s = f_{\text{scene}}(\phi(s)),
\)
which serves as the episodic memory anchor for downstream tasks.

\subsubsection{Retrieval fusion}
\label{sec:rf}
Rather than replacing or re-scoring candidates, we use episodic matches as a lightweight reranking signal.
{A semantically retrieved memory is prioritized if it is also aligns with the current scene and character roles}. 
This preserves broad recall while promoting memories that truly belong to the same event as the present moment.
Specifically, we first perform semantic and episodic retrieval to obtain a ranked list of top-k candidate memory \(\mathcal{E}_{\text{sem}}(q)\) and \(\mathcal{E}_{\text{epi}}(q)\), respectively.
Any memory entry that appears in both the semantic and episodic results is moved to the front of the final list, which preserves their relative order from the semantic ranking. 
Entries that are only in the semantic result retain their original positions. 
Formally, let \(\mathcal{E}_{\text{sem}}(q)=\{m_1,m_2,\dots,m_k\}\) which is ordered by similarity score.
The final fusion retrieval result \(\mathcal{E}(q)\) is constructed as,
\(
\mathcal{E}(q) = \{ m \in \mathcal{E}_{\text{sem}}(q) \mid m \in \mathcal{E}_{\text{epi}}(q) \} \cup \{ m \in \mathcal{E}_{\text{sem}}(q) \mid m \notin \mathcal{E}_{\text{epi}}(q) \}
\),
where the relative order within each subset follows that in $\mathcal{E}_{\text{sem}}(q)$.
Notably, our fusion mechanism does not introduce any new candidates beyond the top-k \(\mathcal{E}_{\text{sem}}(q)\).

\section{Experiments}
\subsection{Settings}
\textbf{Datasets.} Our main comparison covers two benchmarks:
\textbf{LOCOMO}\cite{maharana2024evaluating} and \textbf{epbench}\cite{2025epmembench}.
\textbf{LOCOMO} evaluates long-term conversational memory through multi-session dialogues, each paired with annotated questions including single-hop, multi-hop, temporal, and open-domain types,
which are denoted as \textbf{$\text{locomo\_single},\text{locomo\_multi},\text{locomo\_time}$} and \textbf{$\text{locomo\_open}$} in our experiments, respectively.
Adversarial (unanswerable) questions are excluded due to missing ground-truth answers and the expectation that systems should reject rather than answer them\cite{chhikara2025Mem0}.
\textbf{Epbench}\cite{2025epmembench} is designed to evaluate performance of models across various episodic memory tasks.
Three datasets(default short Claude, sci-fi short Claude and news short Claude) from epbench are considered in our experiments,
which are denoted as \textbf{$\text{ep\_default},\text{ep\_scifi}$} and \textbf{$\text{ep\_news}$}.

\paragraph{Metrics.}
Prior work in conversational memory\cite{goswami2025dissecting,soni2024evaluating,singh2020ensemble} typically uses metrics like F1 and BLEU-1,
which show limitations in conversational contexts when evaluating factual accuracy\cite{chhikara2025Mem0}.
To better assess performance, we follow the method in \cite{chhikara2025Mem0},
which adopt \textbf{LLM-as-a-Judge (J)} and \textbf{F1} as our metric.
Specifically, J denotes that an LLM evaluates responses for factual accuracy, relevance, and completeness by comparing them against ground-truth answers. 
We run 5 times per method and report average values due to stochasticity.

\paragraph{Configurations.} 
All methods share the same generator LLM gpt-4o-mini and embedding model text-embedding-3-small.
In basic configuration, we set the number of retrieval passages $k=5$, the adjacent window size $w=1$.
Meanwhile, parameters about scene aggregation $\delta_t=1, \delta_\tau=0.7$.
In addition, we use Llama-3.3-70B-Instruct\cite{llama3modelcard} to do NER and OpenIE to extract triples and construct semantic graphs.

\paragraph{Baselines.}
We compare against several representative memory approaches. 
\textbf{Traditional RAG} uses a standard dense retriever, which embeds corpus chunks, retrieves the top-$k$ by vector similarity. 
\textbf{HippoRAG}\cite{gutierrez2025rag} as a multihop RAG, which enhances dense retrieval with a graph-based semantic index. 
\textbf{Popular memory plugins} (Zep, Mem0, LangMem) are included too, which maintain persistent dialogue memories (typically as summaries or structured entries) and retrieve a small set of relevant memories at inference.

\subsection{Main results}

\begin{table*}[!th]
\centering
\begin{adjustbox}{width=\textwidth}
\begin{tabular}{c|llllllllllllll}
\toprule
\multirow{2}{*}{\textbf{Method}} & \multicolumn{2}{c}{\textbf{$\text{locomo\_open}$}}  &  \multicolumn{2}{c}{\textbf{$\text{locomo\_multi}$}} &  \multicolumn{2}{c}{$\textbf{locomo\_single}$} &  \multicolumn{2}{c}{\textbf{$\text{locomo\_time}$}} &  \multicolumn{2}{c}{\textbf{$\text{ep\_default}$}}&  \multicolumn{2}{c}{\textbf{$\text{ep\_scifi}$}}&  \multicolumn{2}{c}{\textbf{$\text{ep\_news}$}}\\
&$\text{F}_1$  &J  &$\text{F}_1$  &J &$\text{F}_1$  &J &$\text{F}_1$  &J &$\text{F}_1$  &J &$\text{F}_1$  &J &$\text{F}_1$  &J \\
\bottomrule
\toprule
RAG & 36.87  &56.60  &12.15 &32.29 &20.95 &36.52 &21.63 &31.15 &18.03 &48.68 &16.53 &59.07 &19.18 &54.89\\
LangMem &40.23 &69.54 &25.26 &50.02 &31.45 &30.75 &25.42 &25.42 &35.45 &71.49 &38.60 &73.45 &33.80 &62.67 \\
Mem0 &28.70 &48.87 &28.64 &35.42 & 25.08 &41.49 &\textbf{48.95} & 57.94 &16.59 &28.29 & 13.49 &32.30 &18.82 &36.44 \\
Zep &48.78 &72.89 &26.45 &42.71 &\textbf{35.25} &52.48 &43.15 &51.40 &32.24 &60.31 &34.99 &63.50 &26.95 &44.89 \\
HippoRAG &42.32 &67.78 &25.53 &55.21 &30.40 &54.26 &51.39 &63.86 &53.25 &91.30 &50.43 &94.49 &47.19 &\textbf{92.22} \\
\midrule
CAST &\textbf{62.02} &\textbf{81.21} &\textbf{29.53} &\textbf{58.33} &33.94 &\textbf{54.63} &49.77 &\textbf{68.22} &\textbf{52.60} &\textbf{93.12} &\textbf{52.04} &\textbf{97.06} &\textbf{51.23} &91.48 \\
\bottomrule
\end{tabular}\end{adjustbox}
\caption{Experiment comparison results of memory-enabled systems on different datasets and methods. 
Evaluation metrics include F1 score($\text{F}_1$) and LLM-as-a-Judge score (J). \textbf{Bold} indicates the best performance across all methods.}
\label{tab:result} 
\end{table*}

Main results are shown in Table~\ref{tab:result}, and three key conclusions can be deduced as below.

\paragraph{CAST has outstanding performance overall, with the largest gains on open conversational memory.}
CAST achieves the best performance on locomo\_open ($\text{F}_1$=62.02, J=81.21), outperforming the strongest memory-plugin baseline Zep by +13.24 $\text{F}_1$ and +8.32 J, and surpassing vanilla RAG by
+25.15 $\text{F}_1$ and +24.61 J. 
On epbench, CAST remains highly competitive, which attains the best judge scores on ep\_default and ep\_scifi.

\paragraph{CAST mainly improves episodic correctness.}
CAST’s advantage on LOCOMO is expressed more reliably in J rather than in surface-overlap metrics.
Even when $\text{F}_1$ is not the highest (e.g., locomo\_single), CAST still achieves a strong J (54.63),
which indicates improved episodic correctness via coherent, person-conditioned scenes rather than merely increasing lexical overlap.

\paragraph{CAST improves robustly across diverse queries.}
Mem0 shows a pronounced instability, which performs relatively well on locomo\_time (48.95 $\text{F}_1$, 57.94 J) while dropping substantially on other domains such as locomo\_open (28.70 $\text{F}_1$, 48.87 J) and
locomo\_multi (28.64 $\text{F}_1$, 35.42 J). 
This variance supports our claim that single memory structures are not robust across heterogeneous queries, whereas CAST provides a more stable design for agents.

\subsection{Analysis}
\label{sec:analysis}
\begin{figure*}[!t]
	\centering 
	\includegraphics[width=\linewidth]{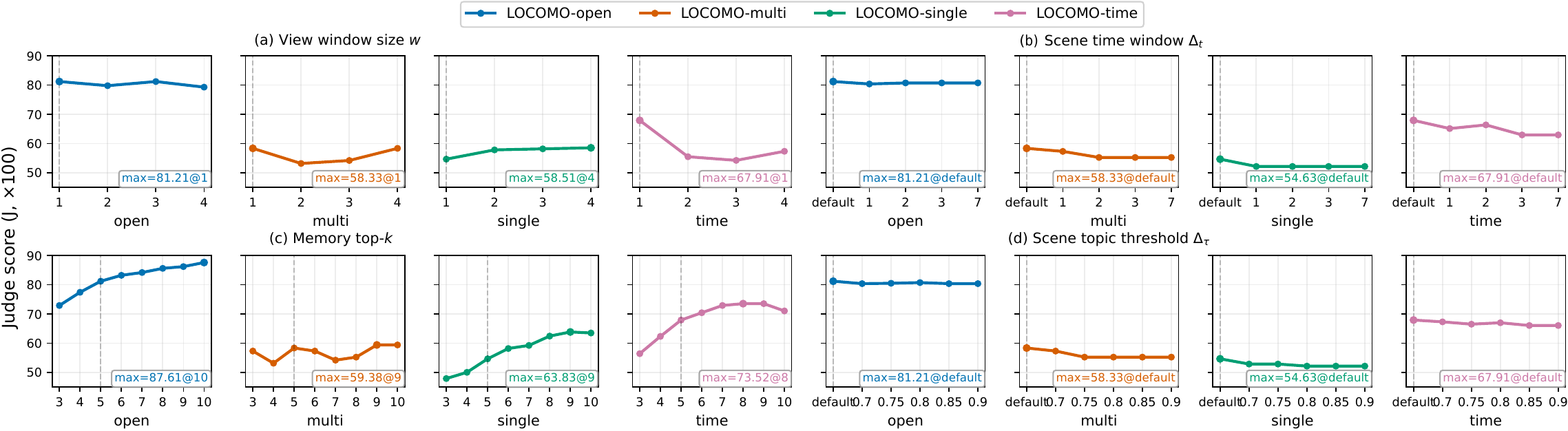}
	\caption{The analysis experiments of CAST. Among them, (a) denotes the results of view parameters, (b) and (d) denote the results of scene aggregation parameters, (c) denotes the results of retrieval parameters.}
	\label{fig:analysis}
\end{figure*}
\subsubsection{Parameters Analysis}
\paragraph{View parameters.}
The window size $w$ is varied from 1 to 5 on the four datasets of LOCOMO.
As in Figure~\ref{fig:analysis}(a),
$w=1$ gives the strongest locomo\_open and locomo\_time scores, while larger windows modestly improve locomo\_single and sometimes locomo\_multi.
Overall, increasing $w$ tends to
dilute temporal alignment (hurting locomo\_time) while providing more local context that can help single-hop recall.

\paragraph{Scene aggregation parameters.} 
We evaluate the two most consequential clustering parameters,
the temporal window \(\Delta_t\) , and the topic similarity threshold \(\Delta_\tau\) on the four datasets of LOCOMO.
As in Figure~\ref{fig:analysis}(b)(d),
Varying \(\Delta_t\) has a mild effect overall, with the best performance at the smallest window. 
Changing \(\Delta_\tau\) yields only modest variation, with the baseline setting performing best overall. 
Higher thresholds (0.75–0.9) generally reduce locomo\_multi and locomo\_single, which indicates
that overly strict topic matching fragments scenes and hurts aggregation.

\paragraph{Retrieval parameters.} 
We vary the number of retrieval evidences $k$ on the four datasets of LOCOMO.
Figure~\ref{fig:analysis}(c) shows that increasing top $k$ generally improves performance, especially on locomo\_open and locomo\_time/locomo\_single, while locomo\_multi benefits less,
suggesting extra memories can introduce distractors for multi-hop queries. 
Overall, $k$=8–9 is the best trade-off range.

\subsubsection{Ablations}
\begin{figure}[!t]
	\centering 
	\includegraphics[width=\columnwidth]{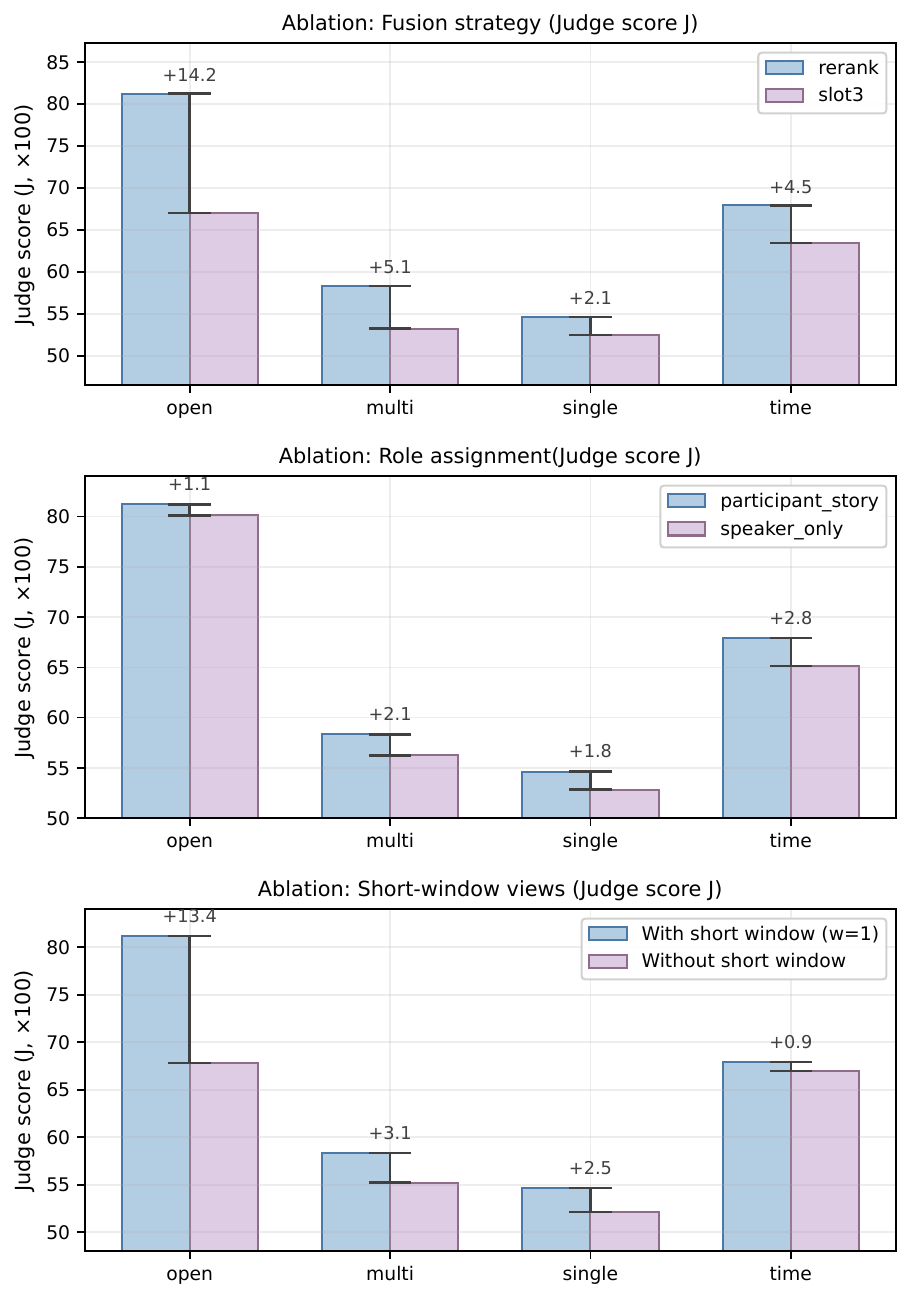}
	\caption{The ablation experiments of CAST.}
	\label{fig:ablation}
\end{figure}

We ablate three parts of CAST on four LOCOMO splits,
(i) short-window view construction, 
(ii) role assignment, 
and (iii) fusion strategy 
as in Figure~\ref{fig:ablation}.

\paragraph{Short-window view construction.} 
Two settings, with window ($w=$1) and without window are compared.
Results indicate that short-window views are critical, especially for locomo\_open.
Removing the adjacent short window causes a large drop on locomo\_open (-13.38 J), while the impact on the other splits is smaller.
This indicates that the short window mitigates QA misalignment across turns and makes the scene more complete.

\paragraph{Role Assignment.} 
We compare indexing using all mentioned persons(participant\_story) versus only the speakers(speaker\_only).
Results show that our role assignment improves robustness, especially on time-sensitive queries.
Actually, in CAST, scenes can be retrieved even when the target person is only mentioned rather than speaking.

\paragraph{Fusion strategy.} 
The fusion strategy of reranking (details in Section~\ref{sec:rf}) and naive slot3 method (inserting a scene retrieved passage at position 3, and preserving the remaining semantic results) are compared.
Results show that slot3 injection hurts across datasets,
which indicates that blindly forcing a memory item into a fixed position may
inject off-topic scenes, finally harmful for answers.

\subsubsection{Use examples}
We illustrate how CAST resolves a common failure mode of semantic-only memory which returns evidence that is topically related but episodically misaligned.
Consider the example below.

\noindent[8 May, \(\ell{=}\)Dance studio] \emph{Caroline}: ``Gina told me her team won first place last night.`` \\
\emph{Melanie}: ``Nice—what did they perform?`` \\
\emph{Caroline}: ``A contemporary piece called \emph{Finding Freedom}.`` \\
Later, the dialogue contains generic dance talk (e.g., ``Gina likes contemporary dance'') without re-stating the piece name.\\
The query is \emph{``What was the name of the piece Gina performed when her team won first place?``}

\noindent\textbf{Semantic-only retrieval (typical).} 
Dense retrieval often gives the later generic ``contemporary dance`` mentions, which are semantically close to the query but do not contain the decisive entity \emph{``Finding Freedom``}. 
This pushes the generator toward vague or hallucinated answers.

\noindent\textbf{CAST.} 
CAST assigns the relevant views to Gina’s stream (as SC), clusters them into a scene under the time/place/topic constraints, and summarizes the scene into a single retrieval unit that preserves the crucial summary ``Gina won the first place in dance``.
At inference time, CAST returns this Gina-conditioned scene, and promotes it within the semantic top-\(k\), which yields a grounded answer with an explicit episodic trace (\emph{who/when/where}).

\section{Conclusions}
We presented CAST, a character-and-scene based memory architecture that complements semantic memory with explicit episodic memory grounded in time, place, and action.
Specifically, CAST constructs character streams, clusters them into coherent scenes, and retrieves them to support downstream tasks.
Experiments have shown that CAST improves most when questions hinge on \emph{who} did \emph{what} \emph{when} and \emph{where}, while remaining competitive when purely semantic cues dominate.

\section*{Limitations}
Despite the promising results, our work has several limitations. 
First, the current approach to defining scene boundaries relies on fixed partitioning strategies, which may fail to capture the dynamic and complex structure of real-world scenarios. 
Second, the fusion mechanism of memories employs a static strategy, lacking the ability to dynamically adjust various demands. 
In addition, the construction of semantic graphs is time-consuming.
These constraints highlight the need for more efficient and adaptive solutions in future work.

\bibliography{custom}

@article{tulving1972episodic,
	title={Episodic and semantic memory},
	author={Tulving, Endel and others},
	journal={Organization of memory},
	volume={1},
	number={381-403},
	pages={1},
	year={1972},
	publisher={New York}
}

@article{schacter1994memory,
  title={Memory systems of 1999.},
  author={Schacter, Daniel L and Wagner, Anthony D and Buckner, Randy L},
  year={2000},
  publisher={Oxford University Press}
}

@article{richards1895butcher,
	title={Butcher on Aristotle's Poetics-Aristotle's Theory of Poetry and Fine Art, by SH Butcher, Litt. D., LL. D. Macmillan. 1895. 10s. net.},
	author={Richards, Herbeht},
	journal={The Classical Review},
	volume={9},
	number={4},
	pages={213--215},
	year={1895},
	publisher={Cambridge University Press}
}

@book{corrigan2012film,
	title={The film experience: An introduction},
	author={Corrigan, Timothy and White, Patricia},
	year={2012},
	publisher={Macmillan}
}

@inproceedings{bamman2013learning,
	title={Learning latent personas of film characters},
	author={Bamman, David and O’Connor, Brendan and Smith, Noah A},
	booktitle={Proceedings of the 51st Annual Meeting of the Association for Computational Linguistics (Volume 1: Long Papers)},
	pages={352--361},
	year={2013}
}

@book{egri1972art,
	title={The art of dramatic writing: Its basis in the creative interpretation of human motives},
	author={Egri, Lajos},
	year={1972},
	publisher={Simon and Schuster}
}

@article{rasmussen2025zep,
	title={Zep: a temporal knowledge graph architecture for agent memory},
	author={Rasmussen, Preston and Paliychuk, Pavlo and Beauvais, Travis and Ryan, Jack and Chalef, Daniel},
	journal={arXiv preprint arXiv:2501.13956},
	year={2025}
}

@article{chhikara2025mem0,
	title={Mem0: Building production-ready ai agents with scalable long-term memory},
	author={Chhikara, Prateek and Khant, Dev and Aryan, Saket and Singh, Taranjeet and Yadav, Deshraj},
	journal={arXiv preprint arXiv:2504.19413},
	year={2025}
}

@article{gutierrez2025rag,
	title={From rag to memory: Non-parametric continual learning for large language models},
	author={Guti{\'e}rrez, Bernal Jim{\'e}nez and Shu, Yiheng and Qi, Weijian and Zhou, Sizhe and Su, Yu},
	journal={arXiv preprint arXiv:2502.14802},
	year={2025}
}

@inproceedings{karpukhin2020dense,
	title={Dense Passage Retrieval for Open-Domain Question Answering.},
	author={Karpukhin, Vladimir and Oguz, Barlas and Min, Sewon and Lewis, Patrick SH and Wu, Ledell and Edunov, Sergey and Chen, Danqi and Yih, Wen-tau},
	booktitle={EMNLP (1)},
	pages={6769--6781},
	year={2020}
}

@inproceedings{park2023generative,
	title={Generative agents: Interactive simulacra of human behavior},
	author={Park, Joon Sung and O'Brien, Joseph and Cai, Carrie Jun and Morris, Meredith Ringel and Liang, Percy and Bernstein, Michael S},
	booktitle={Proceedings of the 36th annual acm symposium on user interface software and technology},
	pages={1--22},
	year={2023}}

@article{zhang2018personalizing,
	title={Personalizing dialogue agents: I have a dog, do you have pets too?},
	author={Zhang, Saizheng and Dinan, Emily and Urbanek, Jack and Szlam, Arthur and Kiela, Douwe and Weston, Jason},
	journal={arXiv preprint arXiv:1801.07243},
	year={2018}
}

@book{aristotle1996poetics,
	title     = {Poetics},
	author    = {Aristotle},
	translator= {Malcolm Heath},
	year      = {1996},
	publisher = {Penguin Classics}
}

@article{beurer2023memgpt,
	title         = {MemGPT: Towards LLMs as Operating Systems},
	author        = {Beurer-Kellner, Lukas and Schmid, Philipp and von Brandenstein, Jonas and others},
	journal       = {arXiv preprint arXiv:2310.08536},
	year          = {2023}
}

@inproceedings{chambers2008unsupervised,
	title     = {Unsupervised Learning of Narrative Event Chains},
	author    = {Chambers, Nathanael and Jurafsky, Daniel},
	booktitle = {Proceedings of the 46th Annual Meeting of the Association for Computational Linguistics (ACL)},
	year      = {2008}
}

@article{conway2000construction,
	title     = {The Construction of Autobiographical Memories in the Self-Memory System},
	author    = {Conway, Martin A. and Pleydell-Pearce, Christopher W.},
	journal   = {Psychological Review},
	volume    = {107},
	number    = {2},
	pages     = {261--288},
	year      = {2000}
}

@inproceedings{guu2020realm,
	title     = {{REALM}: Retrieval-Augmented Language Model Pre-Training},
	author    = {Guu, Kelvin and Lee, Kenton and Tung, Zora and Pasupat, Panupong and Chang, Ming-Wei},
	booktitle = {Proceedings of the 37th International Conference on Machine Learning (ICML)},
	year      = {2020}
}

@article{hearst1997texttiling,
	title     = {TextTiling: Segmenting Text into Multi-paragraph Subtopic Passages},
	author    = {Hearst, Marti A.},
	journal   = {Computational Linguistics},
	volume    = {23},
	number    = {1},
	pages     = {33--64},
	year      = {1997}
}

@inproceedings{izacard2021leveraging,
  title={Leveraging passage retrieval with generative models for open domain question answering},
  author={Izacard, Gautier and Grave, Edouard},
  booktitle={Proceedings of the 16th conference of the european chapter of the association for computational linguistics: main volume},
  pages={874--880},
  year={2021}
}

@inproceedings{maharana2024evaluating,
  title={Evaluating Very Long-Term Conversational Memory of LLM Agents},
  author={Maharana, Adyasha and Lee, Dong-Ho and Tulyakov, Sergey and Bansal, Mohit and Barbieri, Francesco and Fang, Yuwei},
  booktitle={Proceedings of the 62nd Annual Meeting of the Association for Computational Linguistics (Volume 1: Long Papers)},
  pages={13851--13870},
  year={2024}
}

@inproceedings{lewis2020retrieval,
	title     = {Retrieval-Augmented Generation for Knowledge-Intensive NLP Tasks},
	author    = {Lewis, Patrick and Perez, Ethan and Piktus, Aleksandra and others},
	booktitle = {Advances in Neural Information Processing Systems (NeurIPS)},
	year      = {2020}
}

@inproceedings{mildenhall2020nerf,
	title     = {{NeRF}: Representing Scenes as Neural Radiance Fields for View Synthesis},
	author    = {Mildenhall, Ben and Srinivasan, Pratul P. and Tancik, Matthew and Barron, Jonathan T. and Ramamoorthi, Ravi and Ng, Ren},
	booktitle = {Proceedings of the 16th European Conference on Computer Vision (ECCV)},
	year      = {2020}
}

@article{schacter1998cognitive,
  title={The cognitive neuroscience of constructive memory},
  author={Schacter, Daniel L and Norman, Kenneth A and Koutstaal, Wilma},
  journal={False-memory creation in children and adults},
  pages={136--175},
  year={2000},
  publisher={Psychology Press}
}

@book{schank1977scripts,
	title     = {Scripts, Plans, Goals and Understanding: An Inquiry into Human Knowledge Structures},
	author    = {Schank, Roger C. and Abelson, Robert P.},
	year      = {1977},
	publisher = {Lawrence Erlbaum Associates}
}

@inproceedings{seitz2006comparison,
	title     = {A Comparison and Evaluation of Multi-View Stereo Reconstruction Algorithms},
	author    = {Seitz, Steven M. and Curless, Brian and Diebel, James and Scharstein, Daniel and Szeliski, Richard},
	booktitle = {Proceedings of the 2006 IEEE Computer Society Conference on Computer Vision and Pattern Recognition (CVPR)},
	year      = {2006}
}

@inproceedings{sitzmann2019scene,
	title     = {Scene Representation Networks: Continuous 3D-Structure-Aware Neural Scene Representations},
	author    = {Sitzmann, Vincent and Zollh{\"o}fer, Michael and Wetzstein, Gordon},
	booktitle = {Advances in Neural Information Processing Systems (NeurIPS)},
	year      = {2019}
}

@inproceedings{zhong2024memorybank,
  title={Memorybank: Enhancing large language models with long-term memory},
  author={Zhong, Wanjun and Guo, Lianghong and Gao, Qiqi and Ye, He and Wang, Yanlin},
  booktitle={Proceedings of the AAAI Conference on Artificial Intelligence},
  volume={38},
  number={17},
  pages={19724--19731},
  year={2024}
}

@article{zacks2007event,
  title={Event segmentation},
  author={Zacks, Jeffrey M and Swallow, Khena M},
  journal={Current directions in psychological science},
  volume={16},
  number={2},
  pages={80--84},
  year={2007},
  publisher={SAGE Publications Sage CA: Los Angeles, CA}
}

@article{wu2025human,
  title={From human memory to ai memory: A survey on memory mechanisms in the era of llms},
  author={Wu, Yaxiong and Liang, Sheng and Zhang, Chen and Wang, Yichao and Zhang, Yongyue and Guo, Huifeng and Tang, Ruiming and Liu, Yong},
  journal={arXiv preprint arXiv:2504.15965},
  year={2025}
}

@inproceedings{fountas2025human,
  title={Human-inspired episodic memory for infinite context LLMs},
  author={Fountas, Zafeirios and Benfeghoul, Martin and Oomerjee, Adnan and Christopoulou, Fenia and Lampouras, Gerasimos and Ammar, Haitham Bou and Wang, Jun},
  booktitle={The Thirteenth International Conference on Learning Representations},
  year={2025}
}

@article{brom2010episodic,
  title={Episodic memory for human-like agents and human-like agents for episodic memory},
  author={Brom, Cyril and Lukavsk{\`y}, Ji{\v{r}}{\'\i} and Kadlec, Rudolf},
  journal={International Journal of Machine Consciousness},
  volume={2},
  number={02},
  pages={227--244},
  year={2010},
  publisher={World Scientific}
}

@article{shuster2020deploying,
	title={Deploying lifelong open-domain dialogue learning},
	author={Shuster, Kurt and Urbanek, Jack and Dinan, Emily and Szlam, Arthur and Weston, Jason},
	journal={arXiv preprint arXiv:2008.08076},
	year={2020}
}

@article{2025epmembench,
	title={Episodic Memories Generation and Evaluation Benchmark for Large Language Models},
	author={Huet, Alexis and Ben Houidi, Zied and Rossi, Dario},
	journal={International Conference on Learning Representations},
	year={2025}
}

@inproceedings{komeili2022internet,
	title={Internet-augmented dialogue generation},
	author={Komeili, Mojtaba and Shuster, Kurt and Weston, Jason},
	booktitle={Proceedings of the 60th annual meeting of the Association for Computational Linguistics (Volume 1: Long papers)},
	pages={8460--8478},
	year={2022}
}

@inproceedings{petroni2019language,
	title={Language models as knowledge bases?},
	author={Petroni, Fabio and Rockt{\"a}schel, Tim and Riedel, Sebastian and Lewis, Patrick and Bakhtin, Anton and Wu, Yuxiang and Miller, Alexander},
	booktitle={Proceedings of the 2019 conference on empirical methods in natural language processing and the 9th international joint conference on natural language processing (EMNLP-IJCNLP)},
	pages={2463--2473},
	year={2019}
}

@article{goswami2025dissecting,
	title={Dissecting the Metrics: How Different Evaluation Approaches Yield Diverse Results for Conversational AI},
	author={Goswami, Gaurav},
	journal={Authorea Preprints},
	year={2025},
	publisher={Authorea}
}

@inproceedings{soni2024evaluating,
	title={Evaluating Domain Coverage in Low-Resource Generative Chatbots: A Comparative Study of Open-Domain and Closed-Domain Approaches Using BLEU Scores},
	author={Soni, Arpita and Arora, Rajeev and Kumar, Anoop and Panwar, Dheerendra},
	booktitle={2024 International Conference on Electrical Electronics and Computing Technologies (ICEECT)},
	volume={1},
	pages={1--6},
	year={2024},
	organization={IEEE}
}

@inproceedings{singh2020ensemble,
	title={An ensemble approach for extractive text summarization},
	author={Singh, Prabhjot and Chhikara, Prateek and Singh, Jasmeet},
	booktitle={2020 International Conference on Emerging Trends in Information Technology and Engineering (ic-ETITE)},
	pages={1--7},
	year={2020},
	organization={IEEE}
}

@article{llama3modelcard,
	title={Llama 3 Model Card},
	author={AI@Meta},
	year={2024},
	url = {https://github.com/meta-llama/llama3/blob/main/MODEL_CARD.md}
}

@article{weston2014memory,
	title={Memory networks},
	author={Weston, Jason and Chopra, Sumit and Bordes, Antoine},
	journal={arXiv preprint arXiv:1410.3916},
	year={2014}
}

@inproceedings{zhang2017dynamic,
	title={Dynamic key-value memory networks for knowledge tracing},
	author={Zhang, Jiani and Shi, Xingjian and King, Irwin and Yeung, Dit-Yan},
	booktitle={Proceedings of the 26th international conference on World Wide Web},
	pages={765--774},
	year={2017}
}

@article{hu2025memory,
	title={Memory in the Age of AI Agents},
	author={Hu, Yuyang and Liu, Shichun and Yue, Yanwei and Zhang, Guibin and Liu, Boyang and Zhu, Fangyi and Lin, Jiahang and Guo, Honglin and Dou, Shihan and Xi, Zhiheng and others},
	journal={arXiv preprint arXiv:2512.13564},
	year={2025}
}

@article{han2025legomem,
	title={Legomem: Modular procedural memory for multi-agent llm systems for workflow automation},
	author={Han, Dongge and Couturier, Camille and Diaz, Daniel Madrigal and Zhang, Xuchao and R{\"u}hle, Victor and Rajmohan, Saravan},
	journal={arXiv preprint arXiv:2510.04851},
	year={2025}
}

@inproceedings{lee2023prompted,
	title={Prompted LLMs as Chatbot Modules for Long Open-domain Conversation},
	author={Lee, Gibbeum and Hartmann, Volker and Park, Jongho and Papailiopoulos, Dimitris and Lee, Kangwook},
	booktitle={Findings of the Association for Computational Linguistics: ACL 2023},
	pages={4536--4554},
	year={2023}
}

@article{wang2025mirix,
	title={Mirix: Multi-agent memory system for llm-based agents},
	author={Wang, Yu and Chen, Xi},
	journal={arXiv preprint arXiv:2507.07957},
	year={2025}
}

@article{jimenez2024hipporag,
	title={Hipporag: Neurobiologically inspired long-term memory for large language models},
	author={Jimenez Gutierrez, Bernal and Shu, Yiheng and Gu, Yu and Yasunaga, Michihiro and Su, Yu},
	journal={Advances in Neural Information Processing Systems},
	volume={37},
	pages={59532--59569},
	year={2024}
}

@article{bruner1991narrative,
	title={The narrative construction of reality},
	author={Bruner, Jerome},
	journal={Critical inquiry},
	volume={18},
	number={1},
	pages={1--21},
	year={1991},
	publisher={University of Chicago Press}
}

@inproceedings{liu2020character,
	title={A character-centric neural model for automated story generation},
	author={Liu, Danyang and Li, Juntao and Yu, Meng-Hsuan and Huang, Ziming and Liu, Gongshen and Zhao, Dongyan and Yan, Rui},
	booktitle={Proceedings of the AAAI Conference on Artificial Intelligence},
	volume={34},
	number={02},
	pages={1725--1732},
	year={2020}
}

@inproceedings{angeli-etal-2015-leveraging,
    title = "Leveraging Linguistic Structure For Open Domain Information Extraction",
    author = "Angeli, Gabor  and
      Johnson Premkumar, Melvin Jose  and
      Manning, Christopher D.",
    editor = "Zong, Chengqing  and
      Strube, Michael",
    booktitle = "Proceedings of the 53rd Annual Meeting of the Association for Computational Linguistics and the 7th International Joint Conference on Natural Language Processing (Volume 1: Long Papers)",
    month = jul,
    year = "2015",
    address = "Beijing, China",
    publisher = "Association for Computational Linguistics",
    url = "https://aclanthology.org/P15-1034/",
    doi = "10.3115/v1/P15-1034",
    pages = "344--354"
}

@book{lessing1879hamburgische,
  title={Hamburgische dramaturgie},
  author={Lessing, Gotthold Ephraim},
  year={1879},
  publisher={Walter de Gruyter GmbH \& Co KG}
}

@inproceedings{ma2024context,
  title={Context-Driven Index Trimming: A Data Quality Perspective to Enhancing Precision of RALMs},
  author={Ma, Kexin and Jin, Ruochun and Haotian, Wang and Xi, Wang and Chen, Huan and Tang, Yuhua and Wang, Qian},
  booktitle={Findings of the Association for Computational Linguistics: EMNLP 2024},
  pages={4886--4901},
  year={2024}
}

@inproceedings{YaoMJZW25,
  author       = {Limei Yao and
                  Kexin Ma and
                  Ruochun Jin and
                  Haoqi Zheng and
                  Dong Wang},
  title        = {Enhancing Vector Data Quality through Negative Learning for Retrieval-augmented
                  Large Models},
  booktitle    = {International Joint Conference on Neural Networks, {IJCNN} 2025, Rome,
                  Italy, June 30 - July 5, 2025},
  pages        = {1--8},
  publisher    = {{IEEE}},
  year         = {2025},
  url          = {https://doi.org/10.1109/IJCNN64981.2025.11229352},
  doi          = {10.1109/IJCNN64981.2025.11229352},
  bibsource    = {dblp computer science bibliography, https://dblp.org}
}

\appendix
\section{Algorithm}
\label{ap:prompt}
We introduce the algorithms for scene aggregation and retrieval fusion,
which are shown in Algorithm~\ref{alg:scene_aggregation} and \ref{alg:retrieval_fusion}, respectively.

\begin{algorithm}[!t]
  \caption{Greedy scene aggregation.}
  \label{alg:scene_aggregation}
  \KwIn{A character stream \(\mathcal{V}^{(p)}=(v_1,\dots,v_{M_p})\) sorted by time; thresholds \((\Delta_t,\Delta_\ell,\Delta_\tau)\); distances \((d_\ell,d_\tau)\); scene encoder \(f_{\text{scene}}\).}
  \KwOut{Scenes \(\mathcal{S}^{(p)}=(s_1,\dots,s_{K_p})\) and vector \(\mathbf{h}_s\) .}
  \(\mathcal{S}^{(p)} \leftarrow\) new empty list\;
  \For(\tcp*[f]{Greedy assignment in temporal order}){\(v_i \in \mathcal{V}^{(p)}\)}{
    \(placed \leftarrow \textbf{False}\)\;
    \For(\tcp*[f]{Scan existing scenes in creation order}){\(s \in \mathcal{S}^{(p)}\)}{
      \(\delta_t \leftarrow |t(v_i) - t_{\max}(s)|\)\;
      \(\delta_\ell \leftarrow d_\ell(\ell(v_i), \ell(s))\)\;
      \(\delta_\tau \leftarrow d_\tau(\tau(v_i), \tau(s))\)\;
      \(m \leftarrow \mathbb{I}[\delta_t \le \Delta_t] + \mathbb{I}[\delta_\ell \le \Delta_\ell] + \mathbb{I}[\delta_\tau \le \Delta_\tau]\)\;
      \If(\tcp*[f]{Two-of-three continuity}){\(m \ge 2\)}{
        add \(v_i\) to \(s\)\;
        \textbf{UpdateSceneStats}(\(s,v_i\))\;
        \(placed \leftarrow \textbf{True}\)\;
        \textbf{break}\;
      }
    }
    \If(\tcp*[f]{Start a new scene}){\(placed=\textbf{False}\)}{
      \(s_{\text{new}} \leftarrow \{v_i\}\)\;
      append \(s_{\text{new}}\) to \(\mathcal{S}^{(p)}\)\;
    }
  }
  \For(\tcp*[f]{Construct scene vectors}){\(s \in \mathcal{S}^{(p)}\)}{
    \(\phi(s) \leftarrow \textsc{Summarize}(\{x(v): v \in s\})\)\;
    \(\mathbf{h}_s \leftarrow f_{\text{scene}}(\phi(s))\)\;
  }
\end{algorithm}

\begin{algorithm}[!t]
  \caption{Retrieval fusion.}
  \label{alg:retrieval_fusion}
  \KwIn{Query \(q\); semantic retrieval top-\(k\) list \(\mathcal{E}_{\text{sem}}(q)\); episodic retrieval top-\(k\) list \(\mathcal{E}_{\text{epi}}(q)\).}
  \KwOut{Fused ranked list \(\mathcal{E}(q)\) of length \(k\).}
  \(\mathcal{E}(q) \leftarrow\) new empty list\;
  \(I \leftarrow \textsc{Set}(\mathcal{E}_{\text{epi}}(q))\)\;
  \tcp{We keep the semantic candidate set fixed, and only promote episodically consistent entries.}
  \For(\tcp*[f]{Promote intersection; preserve semantic order}){\(m \in \mathcal{E}_{\text{sem}}(q)\)}{
    \If{\(m \in I\)}{
      add \(m\) to \(\mathcal{E}(q)\)\;
    }
  }
  \For(\tcp*[f]{Append the remaining semantic entries}){\(m \in \mathcal{E}_{\text{sem}}(q)\)}{
    \If{\(m \notin I\)}{
      add \(m\) to \(\mathcal{E}(q)\)\;
    }
  }
  \Return \(\mathcal{E}(q)\)\;
\end{algorithm}

\paragraph{Complexity for scene aggregation.}
Our algorithm runs in less than $O(n^2)$ time.
Let $N$ denote the number of original views,
each view $v_i$ is assigned to $a_i$ participants, which yields
$M = sum_{i=1..N} \{a_i\}$ per-person views after replication. 
These views are partitioned into $P$ person streams with sizes ${M_p}$.
Within each character stream, we perform our scene aggregation. 
Let one scene-match decision cost \(O(K)\),
then the best case is each view matches the first scene checked (or each person has \(O(1)\) scenes), which yields \(O(M\cdot K)\approx O(M)\).
In the worst case, each event scans all existing scenes, which yields \(O\!\left(\sum_p M_p^2\cdot K\right)\).
In the extreme where all events fall into one bucket, this becomes \(O(M^2\cdot K)\approx O(M^2)\).

\paragraph{Complexity for retrieval fusion.}
Our algorithm runs in $O(n^2)$ time.
At inference time, we retrieve a semantic top-\(k\) list \(\mathcal{E}_{\text{sem}}(q)=(m_1,\dots,m_k)\) and an episodic top-\(k\) list \(\mathcal{E}_{\text{epi}}(q)\).
Our fusion step does not introduce new candidates: it stably promotes entries in the intersection \(\mathcal{E}_{\text{sem}}(q)\cap \mathcal{E}_{\text{epi}}(q)\) to the front while preserving the original semantic order within each block.
Implementationally, we build a hash-set for membership from \(\mathcal{E}_{\text{epi}}(q)\) in \(O(k)\) time and \(O(k)\) space, then scan \(\mathcal{E}_{\text{sem}}(q)\) to output the promoted block and the remainder in overall \(O(k)\) time.
Thus the fusion overhead per query is \(O(k)\) time and \(O(k)\) auxiliary memory, negligible compared to the underlying retrieval and generation costs.

\section{Experiment}
\subsection{Prompt}
\label{sec:prompt}
To evaluate J score, we adapt prompts from\cite{beurer2023memgpt} as in Table~\ref{tab:prompt-j}.
Prompts utilized for LLMs to generate answers are shown in Table~\ref{tab:prompt-r}.
\\

\noindent\rule{\linewidth}{0.4pt}
\noindent
\textbf{Prompt Template for LLM as a Judge}
\noindent\rule{\linewidth}{0.4pt}

Your task is to label an answer to a question as ``CORRECT'' or ``WRONG''. You will be given the following data: (1) a question (posed by one user to another user), (2) a ‘gold’ (ground truth) answer, (3) a generated answer which you will score as CORRECT/WRONG.

The point of the question is to ask about something one user should know about the other user based on their prior conversations. The gold answer will usually be a concise and short answer that includes the referenced topic, for example:
Question: Do you remember what I got the last time I went to Hawaii?
Gold answer: A shell necklace
The generated answer might be much longer, but you should be generous with your grading – as long as it touches on the same topic as the gold answer, it should be counted as CORRECT.

For time related questions, the gold answer will be a specific date, month, year, etc. The generated answer might be much longer or use relative time references (like ‘last Tuesday’ or ‘next month’), but you should be generous with your grading – as long as it refers to the same date or time period as the gold answer, it should be counted as CORRECT. Even if the format differs (e.g., ‘May 7th’ vs ‘7 May’), consider it CORRECT if it’s the same date.

Now it’s time for the real question:

Question: \{question\}

Gold answer: \{gold\_answer\}

Generated answer: \{generated\_answer\}

First, provide a short (one sentence) explanation of your reasoning, then finish with CORRECT or WRONG. Do NOT include both CORRECT and WRONG in your response, or it will break the evaluation script.

Just return the label CORRECT or WRONG in a json format with the key as ``label''.

\vspace{0.5em}
\noindent\rule{\linewidth}{0.4pt} 

\captionof{table}{Prompt Template for LLM as a Judge.}
\label{tab:prompt-j}
\noindent\rule{\linewidth}{0.4pt}
\noindent
\textbf{Prompt Template for Results Generation}
\noindent\rule{\linewidth}{0.4pt} 

You are an intelligent memory assistant tasked with retrieving accurate information from conversation memories.

\medskip
\noindent\textbf{\# CONTEXT:}

\smallskip
You have access to memories from two speakers in a conversation. These memories contain timestamped information that may be relevant to answering the question.

\medskip
\noindent\textbf{\# INSTRUCTIONS:}

\begin{enumerate}
    \item Carefully analyze all provided memories from both speakers.
    
    \item Pay special attention to the timestamps to determine the answer.
    
    \item If the question asks about a specific event or fact, look for direct evidence in the memories.
    
    \item If the memories contain contradictory information, prioritize the most recent memory.
    
    \item If there is a question about time references (like ``last year'', ``two months ago'', etc.), calculate the actual date based on the memory timestamp. For example, if a memory from 4 May 2022 mentions ``went to India last year,'' then the trip occurred in 2021.
    
    \item Always convert relative time references to specific dates, months, or years. For example, convert ``last year'' to ``2022'' or ``two months ago'' to ``March 2023'' based on the memory timestamp. Ignore the reference while answering the question.
    
    \item Focus only on the content of the memories from both speakers. Do not confuse character names mentioned in memories with the actual users who created those memories.
    
    \item The answer should be less than 5--6 words.
\end{enumerate}

\medskip
\noindent\textbf{\# APPROACH (Think step by step):}

\begin{enumerate}
    \item First, examine all memories that contain information related to the question.
    
    \item Examine the timestamps and content of these memories carefully.
    
    \item Look for explicit mentions of dates, times, locations, or events that answer the question.
    
    \item If the answer requires calculation (e.g., converting relative time references), show your work.
    
    \item Formulate a precise, concise answer based solely on the evidence in the memories.
    
    \item Double-check that your answer directly addresses the question asked.
    
    \item Ensure your final answer is specific and avoids vague time references.
\end{enumerate}

\medskip
\noindent
Memories for user \{speaker\_1\_user\_id\}:

\smallskip
\{speaker\_1\_memories\}

\medskip
\noindent
Memories for user \{speaker\_2\_user\_id\}:

\smallskip
\{speaker\_2\_memories\}

\medskip
\noindent
Question: \{question\}

\medskip
\noindent
Answer:

\vspace{0.8em}
\noindent\rule{\linewidth}{0.4pt} 

\captionof{table}{Prompt Template for Results Generation.}
\label{tab:prompt-r}

\subsection{Baseline}
\label{sec:baseline}
\paragraph{HippoRAG.}
HippoRAG is a memory augmented retrieval framework that organizes long-term interactions into episodic memories indexed by entities. 
It uses a graph-based memory store where nodes represent entities and edges capture co-occurrence or temporal relations. 
During retrieval, it performs personalized PageRank(PPR) over this graph to rank relevant memories based on the query’s entity context.
HippoRAG v2 extends this approach by incorporating phrase nodes with passage nodes, realizing a dense-parse structure.
The baseline used in our experiments is HippoRAG v2, with basic configuration explained before.

\paragraph{Mem0.}
Mem0 is a personalized memory system that stores user-specific information in a key-value memory bank.
It extracts and updates user traits or preferences from interactions using an LLM-based summarizer, then retrieves relevant memories via semantic similarity (e.g., cosine similarity over embeddings) between the current query and stored keys. 
Mem0 has a graph augmented version $\text{Mem0}^*$, while the method used in out experiments is the original method.
We have also compared CAST with $\text{Mem0}^*$, where CAST is still excelling.

\paragraph{Zep}
Zep is a session-aware memory system that structures conversation history into summaries. 
It maintains a rolling window of recent interactions and periodically compresses them into higher-level memory entries (e.g., “The user discussed X with Y on [date]”). 
At query time, Zep retrieves relevant episodes via dense vector search over these summaries and injects them into the prompt to provide context-aware responses.

\paragraph{LangMem}
LangMem is a language-model-driven memory architecture that dynamically summarizes and indexes conversational history into a hierarchical memory store. 
It uses an LLM to extract structured memory entries (e.g., facts, events, preferences) from dialogue turns and organizes them into topic-based clusters. 
Retrieval is performed via semantic search over these summaries, and the most relevant memories are fused into the prompt to condition coherent and contextually grounded responses.

\subsection{Result}
\label{sec:result}
The results of analysis experiments are shown as below.
\begin{table}[!ht]
\centering
\caption{Performance of CAST with varying retrieval top-k values on LOCOMO benchmark.}
\label{tab:topk}
\begin{adjustbox}{width=\linewidth}
\begin{tabular}{c|cccc}
\toprule
\textbf{$k$} & \textbf{locomo\_open} & \textbf{locomo\_multi} & \textbf{locomo\_single} & \textbf{locomo\_time} \\
\midrule
3     & 72.89 & 57.29 & 47.87 & 56.39 \\
4     & 77.41 & 53.12 & 50.00 & 62.31 \\
5     & 81.21 & 58.33 & 54.63 & 67.91 \\
6     & 83.23 & 57.29 & 58.16 & 70.41 \\
7     & 84.19 & 54.17 & 59.22 & 72.90 \\
8     & 85.61 & 55.21 & 62.41 & 73.52 \\
9     & 86.21 & 59.38 & 63.83 & 73.52 \\
10    & 87.61 & 59.38 & 63.48 & 71.03 \\
\bottomrule
\end{tabular}
\end{adjustbox}
\end{table}

\begin{table}[!ht]
\centering
\caption{Performance of CAST with different scene time window sizes (in days) on LOCOMO benchmark.}
\label{tab:time_window}
\begin{adjustbox}{width=\linewidth}
\begin{tabular}{c|cccc}
\toprule
\textbf{$\Delta_t$} & \textbf{locomo\_open} & \textbf{locomo\_multi} & \textbf{locomo\_single} & \textbf{locomo\_time} \\
\midrule
1    & 81.21 & 58.33 & 54.63 & 67.91 \\
2    & 80.38 & 57.29 & 52.13 & 65.11 \\
3    & 80.74 & 55.21 & 52.13 & 66.36 \\
7    & 80.74 & 55.21 & 52.13 & 62.93 \\
\bottomrule
\end{tabular}
\end{adjustbox}
\end{table}

\begin{table}[!ht]
\centering
\caption{Performance of CAST with different scene topic thresholds on LOCOMO benchmark.}
\label{tab:topic_threshold}
\begin{adjustbox}{width=\linewidth}
\begin{tabular}{c|cccc}
\toprule
\textbf{$\Delta_\tau$} & \textbf{locomo\_open} & \textbf{locomo\_multi} & \textbf{locomo\_single} & \textbf{locomo\_time} \\
\midrule
0.7    & 81.21 & 58.33 & 54.63 & 67.91 \\
0.75   & 80.38 & 57.29 & 52.84 & 67.29 \\
0.8    & 80.50 & 55.21 & 52.84 & 66.49 \\
0.85   & 80.74 & 55.21 & 52.13 & 66.98 \\
0.9    & 80.38 & 55.21 & 52.13 & 66.04 \\
\bottomrule
\end{tabular}
\end{adjustbox}
\end{table}

\begin{table}[!ht]
\centering
\caption{Ablation study of CAST components on LOCOMO benchmark.}
\label{tab:ablation}
\begin{adjustbox}{width=\linewidth}
\begin{tabular}{l|cccc}
\toprule
\textbf{Ablation} & \textbf{locomo\_open} & \textbf{locomo\_multi} & \textbf{locomo\_single} & \textbf{locomo\_time} \\
\midrule
with window    & 81.21 & 58.33 & 54.63 & 67.91 \\
without window     & 67.83 & 55.21 & 52.13 & 66.98 \\
participant\_story     & 81.21 & 58.33 & 54.63 & 67.91 \\
speaker\_only    & 80.11 & 56.25 & 52.84 & 65.14 \\
rerank          & 81.21 & 58.33 & 54.63 & 67.91 \\
slot3           & 66.98 & 53.21 & 52.48 & 63.45 \\
\bottomrule
\end{tabular}
\end{adjustbox}
\end{table}

\end{document}